\documentclass[letterpaper]{article} 
\usepackage[draft]{aaai2026}  
\usepackage{times}  
\usepackage{helvet}  
\usepackage{courier}  
\usepackage[hyphens]{url}  
\usepackage{graphicx} 
\usepackage{natbib}  
\usepackage{caption} 
\usepackage{dsfont}
\usepackage{algorithm}
\usepackage{algorithmic}   
\usepackage{url}        
\usepackage{booktabs}      
\usepackage{amsfonts}   
\usepackage{nicefrac}  
\usepackage{microtype}      
\usepackage{xcolor}         
\usepackage{amsmath}
\usepackage{amssymb}
\usepackage{mathtools}
\usepackage{amsthm}
\usepackage{dsfont}
\usepackage{subcaption}
\usepackage{bm}
\usepackage{multirow}
\usepackage{tikz}
\usepackage{multicol}

\usepackage{newfloat}
\usepackage{listings}
\DeclareCaptionStyle{ruled}{labelfont=normalfont,labelsep=colon,strut=off} 
\floatstyle{ruled}
\newfloat{listing}{tb}{lst}{}
\floatname{listing}{Listing}
\theoremstyle{plain}
\newtheorem{theorem}{Theorem}

\theoremstyle{definition}
\newtheorem{definition}[theorem]{Definition}

\newtheorem{example}[theorem]{Example}
\newtheorem{remark}[theorem]{Remark}

\renewcommand{\P}{\mathbb{P}}
\newcommand{\E}{\mathbb{E}}

\newcommand{\D}{\mathcal{D}}
\newcommand{\X}{\mathcal{X}}
\newcommand{\Y}{\mathcal{Y}}

\newcommand{\R}{\mathcal{R}}
\renewcommand{\H}{\mathcal{H}}

\newcommand{\id}[1]{\mathds{1}\offf{#1}}

\newcommand{\Eqref}[1]{Eq.~\eqref{#1}}

\newcommand{\off}[1]{\left[#1\right]}
\newcommand{\offf}[1]{\left\{#1\right\}}

\title{Online Conformal Selection with Accept-to-Reject Changes}
\author{
Kangdao Liu \textsuperscript{\rm 1}\equalcontrib,
Huajun Xi \textsuperscript{\rm 2}\equalcontrib,
Chi-Man Vong\textsuperscript{\rm 1},
Hongxin Wei\textsuperscript{\rm 2}
}

\affiliations{
\textsuperscript{\rm 1}Department of Computer and Information Science, University of Macau \\
\textsuperscript{\rm 2}Department of Statistics and Data Science, Southern University of Science and Technology \\
Correspond to \texttt{weihx@sustech.edu.cn}, \texttt{cmvong@um.edu.mo}
}

\usepackage{bibentry}

\begin{document}

\maketitle

\begin{abstract}
Selecting a subset of promising candidates from a large pool is crucial across various scientific and real-world applications.
\textit{Conformal selection} offers a distribution-free and model-agnostic framework for candidate selection with uncertainty quantification.
While effective in offline settings, its application to online scenarios, where data arrives sequentially, poses challenges.
Notably, conformal selection permits the deselection of previously selected candidates, which is incompatible with applications requiring irreversible selection decisions.
This limitation is particularly evident in resource-intensive sequential processes, such as drug discovery, where advancing a compound to subsequent stages renders reversal impractical.
To address this issue, we extend conformal selection to an online \textit{Accept-to-Reject Changes} (ARC) procedure: non-selected data points can be reconsidered for selection later, and once a candidate is selected, the decision is irreversible.
Specifically, we propose a novel conformal selection method, \textit{Online Conformal Selection with Accept-to-Reject Changes} (dubbed \textbf{OCS-ARC}), which incorporates online Benjamini–Hochberg procedure into the candidate selection process.
We provide theoretical guarantees that OCS-ARC controls the false discovery rate (FDR) at or below the nominal level at any timestep under both i.i.d. and exchangeable data assumptions.
Additionally, we theoretically show that our approach naturally extends to multivariate response settings.
Extensive experiments on synthetic and real-world datasets demonstrate that OCS-ARC significantly improves selection power over the baseline while maintaining valid FDR control across all examined timesteps.

\end{abstract}

\section{Introduction}
Selecting a subset of promising candidates from a large pool is crucial in various scientific and real-world applications, such as recruitment screening \cite{Emily2022}, drug discovery \cite{sheridan2015ecounterscreening}, and foundation model alignment \cite{gui2024conformal}.
For instance, in recruitment, companies identify qualified applicants to enhance organizational performance and outcomes.  
This gives rise to the significance of \textit{conformal selection} \cite{jin2023model, jin2023selection}, a systematic framework for candidate selection with uncertainty quantification.  
Specifically, conformal selection constructs a conformal $p$-value to quantify whether a candidate should be selected and applies the Benjamini-Hochberg (BH) procedure \cite{benjamini1995controlling, benjamini2001control} to control the False Discovery Rate (FDR) with a theoretical guarantee.
Importantly, this framework imposes no parametric assumptions about the data distribution and is applicable to any black-box predictor,  making it a trustworthy and powerful technique for candidate selection.

While conformal selection excels in offline settings, its application to online scenarios, where data arrives sequentially, poses challenges.
In particular, when applied to this setting, conformal selection allows previously selected candidates to be deselected later.
However, such reversibility may be incompatible with some real-world applications:
\begin{itemize}
    \item   In drug discovery, due to high costs and time constraints, compound selection often occurs in sequential stages.  
Once a compound is selected at any timestep, it may advance to subsequent experiments, rendering reversal at later points impractical.
However, non-selected candidates can be reconsidered later, ensuring that significant compounds are not permanently disregarded.

    \item In recruitment screening, applications are received on a rolling basis and evaluated sequentially in batches until all required positions are filled. Once a candidate is identified as qualified and selected, they may advance to subsequent evaluations (e.g., interviews), with withdrawals at later stages being uncommon. Non-selected candidates may be placed on a waitlist for potential reconsideration.

\end{itemize}
In such cases, a selection mechanism that satisfies an Accept-to-Reject Changes (ARC) procedure is required: non-selected data points can be reconsidered for selection in a later stage, but once a candidate is selected,  the decision is rendered final and irreversible.
Consequently, extending conformal selection to this setting is a critical step toward enabling trustworthy candidate selection in online scenarios.

In this work, we extend conformal selection to the online ARC setting.  
We propose a novel method, \textit{Online Conformal Selection with Accept-to-Reject Changes} (\textbf{OCS-ARC}), which satisfies the ARC property.
In particular, OCS-ARC incorporates the online Benjamini-Hochberg  procedure \cite{fischer2024online2} into the selection of conformal $p$-values.
We provide a theoretical guarantee for OCS-ARC, demonstrating that, under the assumption of independent and identically distributed (i.i.d.) data, the method controls the FDR at or below the nominal level at any timestep.  
This guarantee also extends to more general settings where the data satisfies a natural exchangeability condition.
Moreover,  we demonstrate that OCS-ARC can be naturally extended to multivariate response settings. 
In particular, we formally establish that our method maintains valid FDR control in such scenarios, as long as the  non-conformity functions employed by OCS-ARC satisfy the \textit{regional monotonicity} property.

To verify the validity and effectiveness of OCS-ARC, we conduct extensive experiments to evaluate its performance in terms of FDR and power across both synthetic datasets and real-world applications.
For real-world scenarios, we apply OCS-ARC to recruitment screening, drug property prediction, and LLM decision-making within the conformal alignment framework \cite{gui2024conformal}.  
As this is the first work to extend conformal selection to the online ARC setting, no existing baseline is available for direct comparison.  
To address this, we introduce a baseline by adapting the Bonferroni correction to the online ARC setting.  
The results demonstrate that our proposed method significantly improves the selection power over the baseline, while ensuring valid FDR control at each timestep, across all examined scenarios.
For instance, in the candidate screening dataset, OCS-ARC achieves a power of approximately 0.8 at timestep 200, while the competitor achieves only around 0.1.
In practice, our approach is easy to implement with any predictive algorithm, such as deep learning models, and incurs negligible computational costs.

\section{Preliminary}
\paragraph{Problem Setup.}

We study the problem of candidate selection in \textit{online} setting where the test data arrives in a sequential order \cite{foster2008alpha, javanmard2018online, xu2024online, fischer2024online, fischer2024online2}.
Formally, define the feature space $\mathcal{X}\subseteq\mathbb{R}^d$ and response space $\mathcal{Y}\subseteq\mathbb{R}$.
We sample a sequence of \textit{test} data points $\{(\bm{X}_{n+t},Y_{n+t})\}$, $t\in\mathbb{N}^{+}$, from the joint distribution $\mathcal{P}_{\mathcal{X}\mathcal{Y}}$, where the outcomes $\{Y_{n+t}\}$ are unobserved.
At each timestep $t$, we observe a new instance $\bm{X}_{n+t}$ and receive a pre-defined threshold $c_t$.
Our goal is to \textit{identify a subset of indices $\R_t \subseteq [t]$ from the test data set that maximizes the number of test observations $i \in R_t$ satisfying $Y_i > c_i$.}
For notation shorthand, we denote $[t]=\{1,\cdots,t\}$.

To provide a theoretical guarantee, the selection procedure is required to control the FDR \cite{benjamini1995controlling} at each timestep $t$ below a user-specified level $q$.
The FDR is defined as the expected proportion of false discoveries (i.e. $i\in\R_t$ but $Y_i\leq c_i$) among all selected observations:
$$
\mathrm{FDR}_t=\E\off{\frac{|\R_t\cap\H_t^0|}{|\R_t|}},
$$
where $\H_t^0={i\in[t]:,Y_i\leq c_i}$, with the convention that $0/0$ is defined to be equal to $0$ in the expression above.

To evaluate the performance of the selection, we employ the \textit{power}, defined as the expected proportion of desirable observations (i.e. $Y_t>c_t$) that are correctly selected:
$$
\mathrm{Power}_t=\E\off{\frac{|\R_t\cap\H_t^1|}{|\H_t^1|}},
$$
where $\H_t^1=\{i\in[t]:\,Y_i>c_i\}$. 
An ideal selection procedure should maximize $\mathrm{Power}_t$ while controlling the $\mathrm{FDR}_t$ at or below the user-specified nominal level for each timestep $t$.
\paragraph{Conformal Selection.}
\textit{Conformal selection} \cite{jin2023selection} is a distribution-free and model-agnostic selection framework that provides theoretical guarantees for finite-sample FDR control.
To gather statistical evidence for selection, conformal selection partitions a subset of the training set as the calibration set. 
In the following, we denote calibration set as $\D_{\text{cal}}=\{(\bm{X}_i,Y_i)\}_{i=1}^n$ and test set as $\D_{\text{test}}=\{(\bm{X}_{n+t},Y_{n+t})\}_{t=1}^T$. 
Conformal selection addresses the following multiple hypothesis testing problem:
$$
H_t^0:\,Y_{n+t}\leq c_t
\quad\text{vs.}\quad
H_t^1:\,Y_{n+t}>c_t,\quad
\forall t=1,\cdots,m.
$$
Rejecting the null hypothesis $H_t^0$ indicates the $t$-th test sample should be included in the selection set, i.e., $t\in\R$, as its response is deemed to exceed the given threshold $c_t$. 

To construct the selection set $\R$, conformal selection employs \textit{conformal $p$-value}, which builds upon the conformal inference framework \cite{vovk1999machine, vovk2005algorithmic}. 
Given the pre-trained model, we define a \textit{non-conformity score} function $V:\,\X\times\Y\to\mathbb{R}$.
The non-conformity score measures how well a data point $(\bm{X},Y)$ aligns with the calibration data, with a larger non-conformity score suggesting that the point is more likely to be an outlier relative to the calibration set.
For calibration samples, the non-conformity scores are $V_i=V(\bm{X}_i,Y_i)$, and for test samples, they are $\hat{V}_{n+t}=V(\bm{X}_{n+t},c_t)$, where $c_t$ is the pre-defined threshold that replaces the unobserved $Y_{n+t}$.
These scores are then used to compute conformal $p$-values through a rank-based comparison of $\hat{V}_{n+t}$ against the calibration scores $\{V_i\}_{i=1}^n$.
A smaller conformal $p$-value indicates a lower rank of $\hat{V}_{n+t}$ relative to the calibration scores. 
This provides stronger evidence for rejecting $H_t^0$, i.e., selecting the $t$-th test sample.

To determine the final selection set $\mathcal{R}$, conformal selection employs the BH procedure \cite{benjamini1995controlling, benjamini2001control}, a widely adopted approach for controlling the FDR in multiple hypothesis testing, to select conformal $p$-values. 
The BH procedure guarantees that the overall FDR is maintained at or below the user-specified level, expressed as $\mathrm{FDR} \leq q$. 
A comprehensive explanation of the BH procedure is provided in Appendix A.

\paragraph{Accept-to-reject changes procedure.}
In real-world applications, such as recruitment screening and drug discovery, a selection mechanism that adheres to the \textit{Accept-to-Reject Changes} (ARC) procedure is crucial.
\begin{definition}
An Accept-to-Reject Changes procedure is a nested sequence of selection sets:
$$
\R_1\subseteq\R_2\subseteq\cdots\subseteq\R_t\subseteq\cdots,
$$
where $\R_t$ denotes the selection set at timestep $t$.
\end{definition}  

\begin{remark}
While the term `Accept-to-Reject Changes' might superficially suggest a process where selected items can later be rejected, this interpretation is misleading. 
The term originates from the language of hypothesis testing, where ``rejecting" a null hypothesis corresponds to selecting a data point.
Thus, `Accept-to-Reject Changes' refers to the process that permits transitions from acceptance (non-selection) to rejection (selection), while  prohibiting the reverse.
\end{remark}

An ARC procedure allows non-selected data points to be reconsidered for selection later, and once a point is selected, the decision is final and irreversible.
However, conformal selection is not an ARC procedure: when applying it at different timesteps, it would change its earlier selection into deselection. 
This can be illustrated by the following example:

\begin{example}
Consider a nominal FDR level of $q = 0.1$, with conformal $p$-values given as $p_1 = 0.1$ and $p_2 = 0.2$. Applying the BH procedure at timestep 1 results in the selection set $\mathcal{R}_1 = \{1\}$. At the second timestep, reapplying the procedure yields the selection set $\mathcal{R}_2 = \emptyset$. 
This example illustrates that conformal selection may alter the status of a candidate (e.g., data point 1) from being selected to being deselected.
\end{example}
To empirically validate this observation, we conduct synthetic experiments across various models and record the number of data points that transition from being selected to deselected (i.e., reject-to-accept). 
Detailed setups are provided in Appendix B.
The results in Figure~\ref{fig:motivation} confirm that this framework violates the ARC property.
This motivates our method, which extends conformal selection to an online ARC setting.

\begin{figure}[t]
\centering
\includegraphics[width=0.41\textwidth]{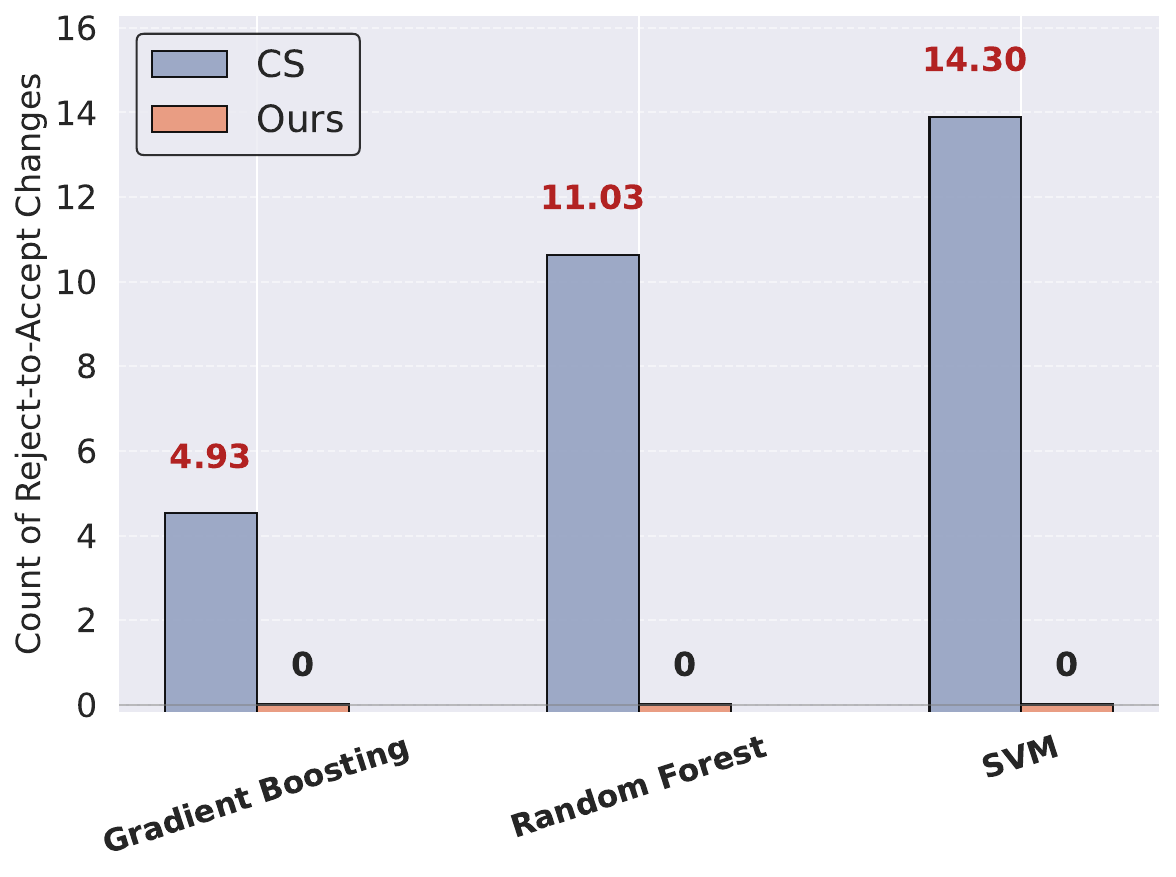}
\caption{
\textbf{Comparison of conformal selection (CS) with OCS-ARC.} 
This experiment measures the total number of reject-to-accept samples for both methods on 500 sequentially arriving data points from a synthetic dataset across three models. Results are averaged over 30 independent runs.
}
  \label{fig:motivation}
\end{figure}

\section{Method}
In this section, we introduce our method, \textit{Online Conformal Selection with Accept-to-Reject Changes} (OCS-ARC). Our approach is composed of three primary steps:
\begin{enumerate}
\item \textbf{Training}: 
Train a predictive model $\hat{\mu}$ for  $Y$. The model can be constructed using any appropriate algorithm.

\item \textbf{Calibration}:
Construct a monotone non-conformity function using the model $\hat{\mu}$ and evaluate it on both the calibration set and the test set. Using the resulting scores, calculate the conformal $p$-values for each test sample.

\item \textbf{Thresholding}:
At each timestep $t$, apply the online BH procedure \cite{fischer2024online2} to the set of conformal $p$-values, yielding the  selection set $\R_t$.
\end{enumerate}
Both the training and calibration steps rely on the labeled set $\D_{\text{train}}$.
In the case where a predictive model $\hat{\mu}$ is already available, our OCS-ARC can be directly applied using the entire labeled set $\D_{\text{train}}$ for calibration.
Otherwise, the training data is split into two subsets: one for model training and the other for calibration. 
For simplicity, in the following discussion, we assume that the model $\hat{\mu}$ is already available (pre-trained), and all labeled data are utilized for calibration. We denote the calibration set as $\D_{\text{cal}} = \{(\bm{X}_i, Y_i)\}_{i=1}^n$ and the test set as $\D_{\text{test}} = \{(\bm{X}_{n+t}, Y_{n+t})\}_{t=1}^T$.  
It is important to emphasize that, in our setting, \textit{the calibration set is fully available beforehand, whereas the test data arrive sequentially in an online manner}. This setup is commonly adopted in prior works \cite{bao2024cap, sale2025online}.

We next provide a detailed explanation of the \textit{calibration} process.
Specifically, we employ conformal $p$-values \citep{bates2023testing, jin2023selection} to perform the hypothesis testing.
When the true responses $\{Y_{n+t}\}_{t=1}^T$ are observed, we compute the non-conformity score $V_{n+t} = V(\bm{X}_{n+t}, Y_{n+t})$. Subsequently, the \textit{oracle} conformal $p$-value is defined as  
$$
p_t^{*}=\frac{\sum_{i=1}^n\id{V_i<V_{n+t}}+U_t(1+\sum_{i=1}^n\id{V_i=V_{n+t}})}{n+1},
$$
where $U_t\sim\mathrm{Uniform}[0,1]$ are i.i.d uniform random variable to randomize over ties when $V_{n+t}$ equals some $V_i$, ensuring a continuous conformal $p$-value.
Standard results from conformal inference ensure that if the data sample $(\bm{X}_{n+t},Y_{n+t})$ follows the same distribution as the calibration data, then the oracle conformal $p$-value $p_t^{*}$ stochastically dominates the uniform distribution on $[0,1]$ \citep{bates2023testing}:
\begin{equation}\label{eq:oracle_conformal_p_uniform}
\P\offf{p_t^{*}\leq\alpha}\leq\alpha,\quad\forall\alpha\in[0,1].
\end{equation}
This property is essential to ensure FDR control.
However, the computation of the oracle conformal $p$-value is \textit{infeasible}, as the true responses $\{Y_t\}_{j=1}^m$ are unobserved.
Therefore, we replace $V_{n+t}$ with its estimate $\hat{V}_{n+t} = V(\bm{X}_{n+t}, c_t)$, resulting in the (practical) conformal $p$-values:
\begin{equation}\label{eq:practical_conformal_p_value}
p_t=\frac{\sum_{i=1}^n\mathds{1}\{V_i<\hat{V}_{n+t}\}+U_t(1+\sum_{i=1}^n\mathds{1}\{V_i=\hat{V}_{n+t}\})}{n+1}.
\end{equation}
Here, $p_j$ quantifies how extreme the threshold is compared to the usual behavior of the outcomes, with a smaller $p_j$ providing stronger evidence for rejecting $H_j^0$.
To ensure that the practical conformal $p$-value $p_j$ stochastically dominates the uniform distribution on $[0,1]$ (see \Eqref{eq:oracle_conformal_p_uniform}), the non-conformity function $V$ is required to satisfy the following \textit{monotonicity} property \citep{jin2023model, jin2023selection}:

\begin{definition}
A non-conformity score function $V(\cdot,\cdot):\,\X\times\Y\to\mathbb{R}$ is monotone if $V(\bm{X},y)\leq V(\bm{X},y^{'})$ holds for any $\bm{X}\in\X$ and any $y,y^{'}\in\Y$ obeying $y\leq y^{'}$.
\end{definition}

\begin{algorithm}[t]
\caption{Online Conformal Selection with Accept-to-Reject Changes (OCS-ARC)}
\label{alg:online_cs_arc}
\begin{algorithmic}[1]
\REQUIRE 
Calibration data $\D_{\text{cal}}=\{(\bm{X}_i,Y_i)\}_{i=1}^n$, 
test data $\D_{\text{test}}=\{(\bm{X}_{n+t},Y_{n+t})\}_{t=1}^T$, 
predefined threshold $\{c_t\}_{t=1}^T$, 
FDR nominal level $q\in(0,1)$,
monotone non-conformity score $V:\,\X\times\Y\to\mathbb{R}$,
real numbers $\{\gamma_t\}_{t=1}^T$.
\STATE Compute $V_i=V(\bm{X}_i,Y_i)$ for $i=1,\cdots,n$.
\FOR{$t=1,\cdots,T$}
\STATE Compute $\hat{V}_{n+t}=V(\bm{X}_{n+t},c_t)$.
\STATE Construct conformal $p$-value $p_t$ as in \Eqref{eq:practical_conformal_p_value}
\STATE Compute online BH procedure threshold:
$$
k^{*}_t=\max\{k\in[t]:\,\sum_{j=1}^t\id{p_j\leq k\alpha\gamma_j}\geq k\}
$$
\STATE Construct selection set:
$
\mathcal{R}_t=\offf{j\in[t]:\, p_j\leq k_t^{*}q\gamma_j}
$
\ENDFOR
\end{algorithmic}
\end{algorithm}

In the Experiments section, we will showcase several examples of monotone non-conformity score functions. After calculating the practical conformal $p$-values, we apply the online BH procedure \citep{fischer2024online2}, which leverages a sequence of real numbers $\{\gamma_t\}_{t=1}^T$ satisfying $\sum_{t=1}^T \gamma_t \leq 1$ for all $T \in \mathbb{N}^{+}$. The complete OCS-ARC procedure is formally outlined in Algorithm~\ref{alg:online_cs_arc}.

\paragraph{Theoretical guarantee.}
We now provide theoretical guarantees for OCS-ARC. First, in the following theorem, we establish that OCS-ARC satisfies the ARC procedure. 

\begin{theorem}
OCS-ARC is an ARC procedure. In particular, the selection sets $\{\R\}_{t=1}^T$ outputted by Algorithm~\ref{alg:online_cs_arc} satisfy
$$
\R_1\subseteq\R_2\subseteq\cdots\subseteq\R_T.
$$
\end{theorem}

The proof is provided in Appendix C.1.  
To validate our theoretical findings, we conduct synthetic experiments on various regression models and count the number of points transitioning from selected to deselected (reject-to-accept). Detailed descriptions of the experimental setups are provided in Appendix B.  
The results, shown in Figure~\ref{fig:motivation}, confirm that OCS-ARC satisfies the properties of an online ARC procedure. Then, we show that Algorithm~\ref{alg:online_cs_arc} controls the FDR at any timestep $t$, under the assumption that the calibration and test data are identically and independently distributed.
\begin{theorem}
\label{the:exc}
Consider a monotone non-conformity score function $V:\,\X\times\Y\to\mathbb{R}$. 
Assume that the calibration data $\{(\bm{X}_i,Y_i)\}_{i=1}^n$ and the test data $\{(\bm{X}_{n+t},Y_{n+t})\}_{t=1}^T$ are independently and identically distributed.
Then, for any nominal level $q\in(0,1)$ and timestep $t$, the selection set $\R_t$ constructed by Algorithm~\ref{alg:online_cs_arc} satisfies $\mathrm{FDR}_t\leq q$.
\end{theorem}

The detailed proof of Theorem \ref{the:exc} is provided in Appendix C.2.
In the following theorem, we extend our previous result to the scenario where the calibration and test samples satisfy a natural exchangeability condition. 
The corresponding technical proof is provided in Appendix C.3.

\begin{theorem}
Consider a monotone non-conformity score function $V$.
Assume that $\{V_1,\cdots,V_n,V_{n+t}\}$ are exchangeable conditional on $\{\hat{V}_{n+t^{'}}\}_{t^{'}=1,t^{'}\neq t}^T$, and $\{V_{t^{'}}\}_{t^{'}=1}^{n+T}$ have no ties almost surely.
Then, for any nominal level $q\in(0,1)$ and timestep $t$, the corresponding selection set $\R_t$ constructed by our OCS-ARC in Algorithm~\ref{alg:online_cs_arc} satisfies $\mathrm{FDR}_t\leq q$.
\end{theorem}

\begin{remark}

It is important to emphasize that our method guarantees FDR control only under the i.i.d. or exchangeability assumption. This guarantee does not hold in dynamic environments with distribution shifts, as seen in online learning.  
The key distinction lies in the availability of feedback. In online learning, after making a decision, the model receives feedback and adapts its policy accordingly. In contrast, our setting lacks such feedback: once a selection is made, no information about its correctness is provided.
\end{remark}

\begin{remark}

We highlight two key theoretical novelties in OCS-ARC.
In standard online BH procedure, the $p$-value are assumed to satisfy conditional validity:
$$
\P\offf{p_t\leq\alpha\mid t\in\H_t^0}\leq\alpha.
$$
However, conformal $p$-value only ensures marginal validity:
$$
\P\offf{p_t\leq\alpha,\ t\in\H_t^0}\leq\alpha.
$$
Moreover, online BH procedure relies on the assumption that $p$-values are \textit{positive regression dependence on a subset} (PRDS) to guarantee FDR control.
Nevertheless, conformal $p$-value fails to meet this requirement.
Due to these issues, directly applying online BH to conformal $p$-value is nontrivial.
    
\end{remark}

\paragraph{Extension to multivariate conformal selection.}


Recent work \cite{bai2025multivariate} extends conformal selection to multivariate response settings. In this section, we show that our OCS-ARC procedure can also be applied in this context.  
Specifically, for a $d^{'}$-dimensional multivariate response $\bm{Y}_{n+t}$, consider the following hypothesis testing:
$$
H_t^0:\,\bm{Y}_{n+t}\in R^{c}
\quad\text{vs.}\quad
H_t^1:\,\bm{Y}_{n+t}\in R,\quad
\forall t=1,\cdots,m,
$$
where $R$ represents an arbitrary pre-defined closed target region in $\mathbb{R}^{d^{'}}$.
To construct a selection set, we define a new non-conformity score function $V(\cdot,\cdot):\,\X\times\Y\to\mathbb{R}$.
We compute $V_i=V(\bm{X}_i,\bm{Y}_i)$, for $i=1,\cdots,n$, and $\hat{V}_{n+t}=V(\bm{X}_{n+t},\bm{r}_t)$, for $t=1,\cdots,T$, where $\bm{r}_t$ is an arbitrary closen point in target region $R$.
Then, we compute the conformal $p$-values defined in \Eqref{eq:practical_conformal_p_value}.
To guarantee that the conformal $p$-value $p_t$ stochastically dominates the uniform distribution on $[0,1]$, the non-conformity function $V$ must satisfy \textit{regional monotonicity} property \citep{bai2025multivariate}:
\begin{definition}
A non-conformity score function $V(\cdot,\cdot):\,\X\times\Y\to\mathbb{R}$ is regional monotone if $V(X,y)\leq V(X,y^{'})$ holds for any $X\in\X$ and any $y\in R^{c}$, $y^{'}\in R$.
\end{definition}
Examples of regional monotone non-conformity score functions are provided in Appendix D.
After computing the conformal $p$-values, the online BH procedure \cite{fischer2024online2} is applied to obtain the final selection set. 
The whole procedure for \textit{Multivariate Online Conformal Selection with Accept-to-Reject Changes} (dubbed mOCS-ARC) is summarized in Appendix D.
In the following theorem, we show that mOCS-ARC controls FDR at any timestep $t$.

\begin{theorem}
\label{the:exc_multi}
Consider a regional monotone non-conformity score function $V:\,\X\times\Y\to\mathbb{R}$.
Assume that the calibration data $\{(X_i,Y_i)\}_{i=1}^n$ and the test data $\{(X_{n+t},Y_{n+t})\}_{t=1}^T$ are independently and identically distributed.
Then, for any nominal level $q\in(0,1)$ and timestep $t$, the selection set $\R_t$ constructed by mOCS-ARC satisfies $\mathrm{FDR}_t\leq q$.
\end{theorem}
The technical proof of this theorem is provided in Appendix C.4. 
This theoretical result establishes that our approach naturally extends to the multivariate response setting.

\begin{figure*}
    \centering
    \includegraphics[width=0.95\textwidth]{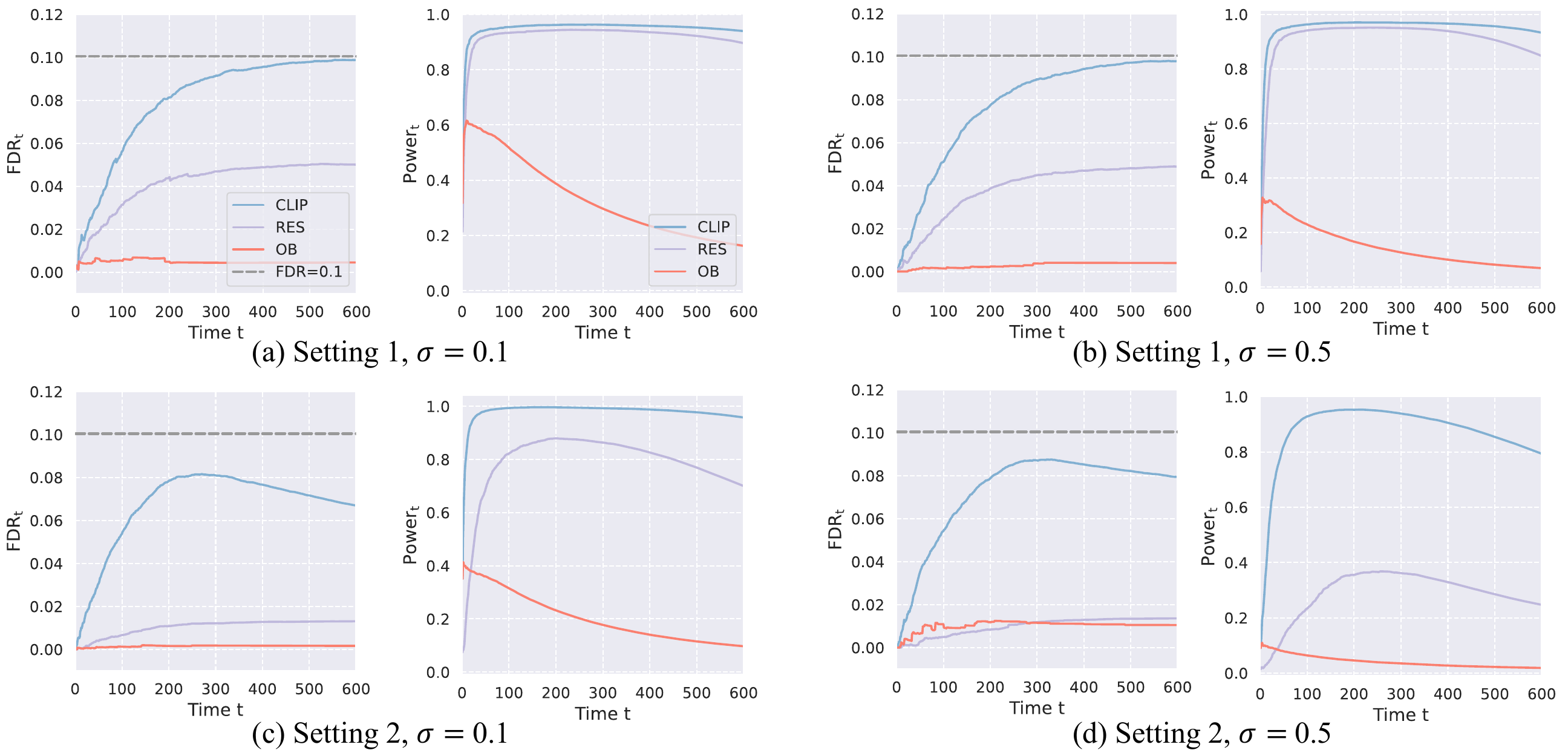} 
    \caption{\textbf{Synthetic data experiments with varying noise levels across different data-generating processes.} The target FDR is set to 0.10, and gradient boosting is used to fit the regression model across all configurations. Subplots (a) and (b) present results for simulation setting 1 (CLIP, RES, and OB, respectively), while subplots (c) and (d) correspond to simulation setting 2.}
    \label{fig:simulation} 
\end{figure*}

\section{Experiments}\label{section:experiment}
We conduct a comprehensive evaluation of  OCS-ARC in terms of FDR and power across synthetic datasets and real-world applications.
For simulated experiments, we follow the setup in prior work \cite{jin2023selection}.
In real-world scenarios, we apply OCS-ARC to candidate screening, drug property prediction, and LLM decision-making within the conformal alignment framework \cite{gui2024conformal}.
\textit{Notably, the experimental setups and results for drug property prediction, along with the relevant mOCS-ARC experiments, are detailed in Appendices F.2 and D, respectively.}

\subsection{Numerical Experiments}\label{section:numerical_exp}
\paragraph{Setup.}
We evaluate our method in simulated datasets for regression.
The task is to select individuals with responses \(Y_{n+t} > 0\) from sequentially arrived test data while controlling FDR at each timestep $t$.
We generate i.i.d. covariates \(\bm{X}_i \sim \text{Uniform}[-1,1]^{20}\) and responses \(\bm{Y}_i = \mu(\bm{X}_i) + \epsilon_i\), where \(\mu(\bm{x}) = \mathbb{E}[\bm{Y} \mid \bm{X} = \bm{x}]\) is nonlinear in \(\bm{x}\), and \(\epsilon_i\) represents random noise independent of \(\bm{X}\). 
To demonstrate the performance of our methods under different data-generating processes, we apply two  simulation settings:
\begin{itemize}
    \item \textbf{Setting 1:} \(\mu(\bm{x}) = 4x_1 \cdot \mathds{1}(x_2 > 0) \cdot \max(0.5, x_3) + 4x_1 \cdot \mathds{1}(x_2 \le 0) \cdot \min(-0.5, x_3)\)
    \item \textbf{Setting 2:} \(\mu(\bm{x}) = 5x_1x_2 + e^{x_4 - 1}\).
\end{itemize}
In both settings, the noise is modeled as \(\epsilon_i \sim N(0, \sigma^2)\) with homogeneous variance. We adjust \(\sigma\) to different levels to generate data with varying noise levels. 
The sizes of the training and calibration datasets are fixed at \(|\mathcal{D}_{\text{train}}| = |\mathcal{D}_{\text{calib}}| = 1000\), while the size of the test dataset is set to \(|\mathcal{D}_{\text{test}}| = 600\).

We utilize gradient boosting and support vector machines (SVM) with a radial basis function kernel to fit a regression model, using the scikit-learn Python library. 
We apply Algorithm~\ref{alg:online_cs_arc} with two different monotone non-conformity score functions, following prior work \cite{jin2023selection}:
\begin{itemize}
    \item $\mathtt{CLIP}$: \(V(x,y) = 1000 \cdot \mathds{1}\{y > 0\} - \hat{\mu}(x)\)
    \item $\mathtt{RES}$: \(V(x,y) = y - \hat{\mu}(x)\).
\end{itemize}
For brevity, we refer to OCS-ARC with $\mathtt{CLIP}$ as CLIP, and OCS-ARC with $\mathtt{RES}$ as RES.
For \(\{\gamma_t\}_{t=1}^T\), we set 
\[
\gamma_{t} = r^{t} \cdot \frac{1 - r}{r},
\]
where \(r\) represents a predefined decay coefficient. 
In our main experiments, \(r\) is fixed at 0.99 without additional tuning. 
Notably, our method exhibits robust performance across a range of $r$ values, as illustrated in Figure \ref{fig:decay_coefficient}.

\paragraph{Baseline.}To the best of our knowledge, this work is the first to extend conformal selection to the online ARC setting, for which \textit{no prior baseline exists}.
To establish a comparison, we introduce a baseline by adapting the Bonferroni correction to the online ARC setting, referred to as \textit{Online Bonferroni correction with Accept-to-Reject Changes} (dubbed OB).
\begin{center}
    OB: Select all $p_t\leq q\gamma_t$ with score function $\mathtt{CLIP}$.
\end{center}
OB employs the same \(\{\gamma_t\}_{t=1}^T\) as in OCS-ARC.
We show that OB satisfies an ARC procedure in Appendix E.

\paragraph{Results.}
The results for gradient boosting are presented in Figure \ref{fig:simulation}, while the results for SVM are included in the Appendix F.1. 
For the evaluation of FDR\(_t\), we average the FDP\(_t\) over 300 independent runs at different timesteps. 
The results show that the proposed method achieves an FDR\(_t\) close to the nominal level 0.1, whereas the baseline exhibits a larger deviation from it.
Among the two non-conformity score functions, $\mathtt{CLIP}$ consistently demonstrates a higher realized FDR\(_t\) across all settings, remaining closer to 0.1. 
Additionally, the FDR\(_t\) stays below 0.1 in all configurations.
For Power\(_t\), we evaluate it by calculating the average proportion of correct selections among all positive samples up to timestep \( t \) across all replicates.
We observe that our methods consistently outperform the baseline OB under all scores and simulation settings.
Moreover, CLIP consistently achieves higher Power\(_t\); however, the power decreases as noise strength increases across all settings. 
This decline arises because increased noise impairs the model accuracy, thereby deteriorating the selection performance.
The results demonstrate that OCS-ARC surpasses the baseline, significantly improving selection power while maintaining valid FDR control at each timestep $t$.

\subsection{Experiments on Recruitment Screening}
\begin{figure}
    \centering
    \begin{subfigure}[b]{0.234\textwidth} 
\includegraphics[width=\linewidth]{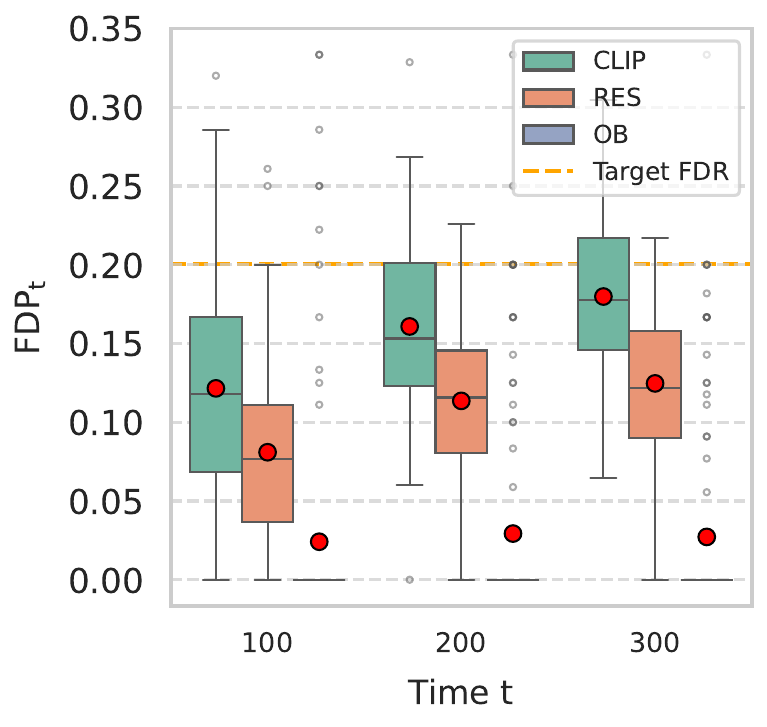}
        \caption{FDP}
    \end{subfigure}
    \hfill 
    \begin{subfigure}[b]{0.234\textwidth}
        \includegraphics[width=\linewidth]{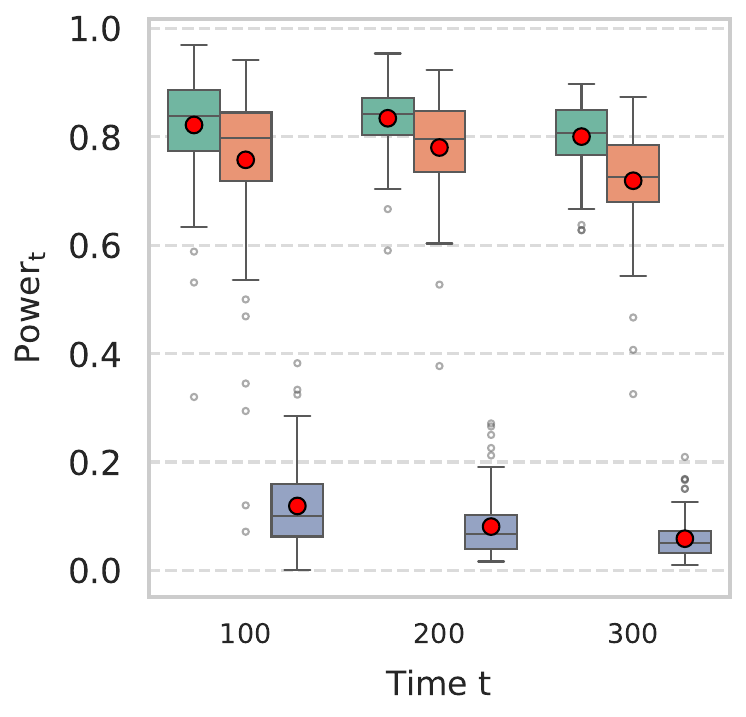}
        \caption{Power}
    \end{subfigure}
\caption{\textbf{Results for candidate screening in recruitment.} The target FDR is set to 0.20, and gradient boosting is used to fit the classifier. FDP$_t$ and Power$_t$ are reported over 100 independent runs at timesteps 100, 200 and 300.}
    \label{fig:candidate}
\end{figure}

We apply OCS-ARC as an automated screening tool for recruitment, where HR staff use machine learning predictions to sequentially evaluate incoming candidates.
Applicants may be either shortlisted for subsequent tests and interviews or placed on a waiting list.
Here, the model estimates the likelihood that a candidate is qualified, with higher scores indicating a better fit.
We employ OCS-ARC to calibrate predictions and produce a shortlist with rigorous FDR control.

\paragraph{Setup.} We utilize a recruitment dataset available on Kaggle \cite{el_kharoua_2024}, which contains 1,500 entries. We divide the dataset into three parts: 700 samples for  training, 400 for calibration, and the remaining 400 for testing. A gradient boosting classifier is employed to perform predictions for this task. 
We aim to select as many data points with labels of 1 as possible while controlling the FDR rate at a predefined level.
Furthermore, the target FDR is set to 0.2 for this experiment, and CLIP is applied as the non-conformity score function.
 
\paragraph{Results.} 
In Figure \ref{fig:candidate}, we use FDP\(_t\) at \(t\) = 100, 200, 300 over 100 replications to evaluate FDR\(_t\). 
The results show that the proposed method achieves an FDR\(_t\) close to the nominal level 0.2, while OB exhibits a large deviation.
For Power\(_t\), we average the proportion of correct selections among all positive samples up to the corresponding timesteps over all replicates.
We observe that our methods consistently outperform OB across all scores.
Overall, the proposed method significantly improves the selection power over the baseline, while ensuring valid FDR control at each timestep.

\begin{figure*}[t]
    \centering
    \begin{subfigure}[b]{0.236\textwidth} 
        \includegraphics[width=\linewidth]{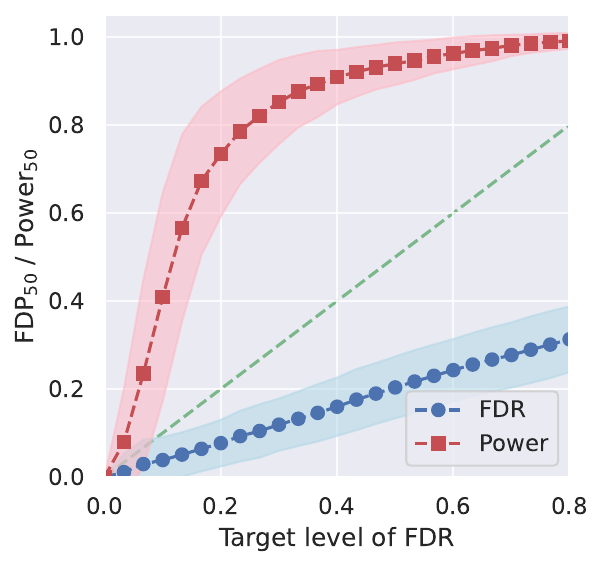}
        \caption{t = 50}
    \end{subfigure}
    \hfill 
    \begin{subfigure}[b]{0.236\textwidth}
        \includegraphics[width=\linewidth]{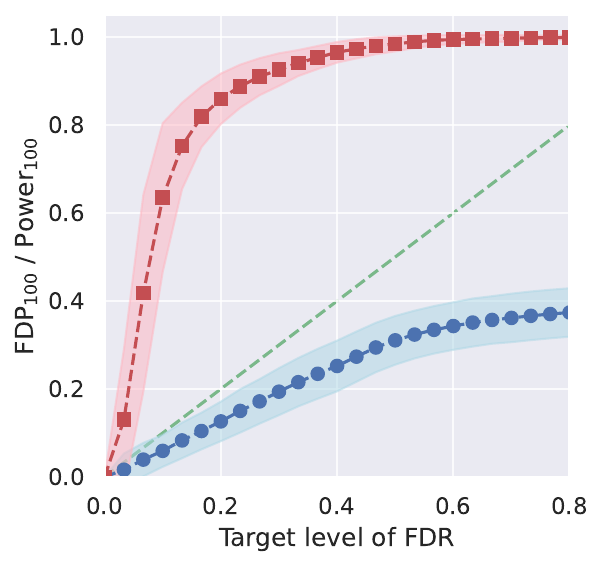}
        \caption{t = 100}
    \end{subfigure}
    \hfill 
    \begin{subfigure}[b]{0.236\textwidth} 
        \includegraphics[width=\linewidth]{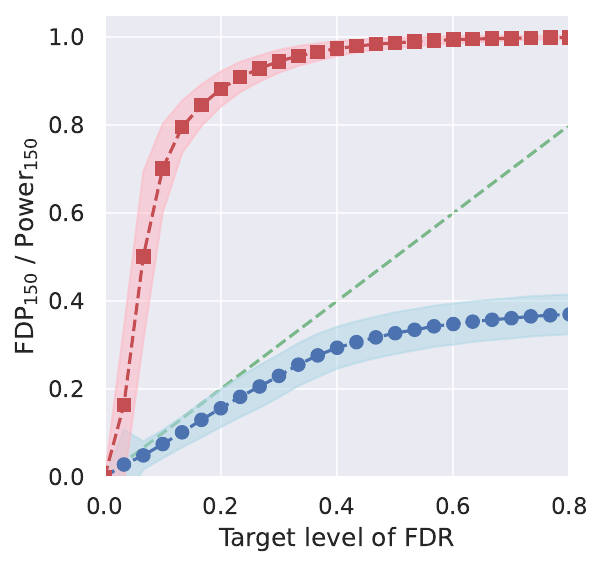}
        \caption{t = 150}
    \end{subfigure}
    \hfill 
    \begin{subfigure}[b]{0.236\textwidth}
        \includegraphics[width=\linewidth]{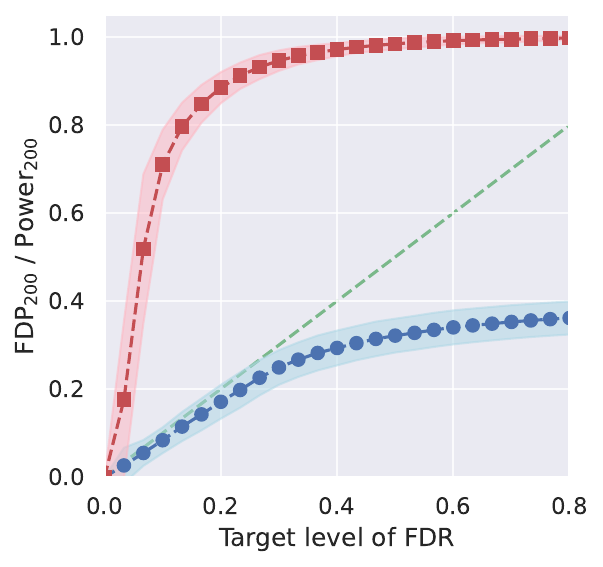}
        \caption{t = 200}
    \end{subfigure}
\caption{\textbf{Results for the application of OCS-ARC to question answering with LLMs} on the TriviaQA Dataset. Experiments are conducted using LLaMA-2-13B-chat, with the target FDR varied from 0 to 0.80 in fixed steps over 25 increments. FDP$_t$ and Power$_t$ are reported over 100 independent runs at timesteps 50, 100, 150, and 200.}
    \label{fig:llmtri}
\end{figure*}

\subsection{Application to Large Language Models}
Given the widespread application of LLMs in decision-making, we show that OCS-ARC can be integrated with LLMs within the conformal alignment framework \cite{gui2024conformal}.
In this experiment, we aim to assess a stream of question-answerings and select high-quality answers while controlling the FDR.
Further details on the conformal alignment framework are provided in Appendix G.
Notably, this procedure can be readily extended to \textit{any} other online candidate selection task utilizing LLMs as base models.

\paragraph{Setup.}
The experiment follows the default settings of conformal alignment \cite{gui2024conformal}, employing a logistic regression model as the alignment score predictor. 
We utilize two widely recognized question-answering datasets: TriviaQA \cite{joshi2017triviaqa} and CoQA \cite{reddy2019coqa}. 
We employ self-evaluation likelihood, input uncertainty scores, and output confidence scores as features to train the alignment score predictor. 
The experiments are conducted using LLaMA-2-13B-chat \cite{touvron2023llama}, with the target FDR varied from 0 to 0.80 in fixed increments over 25 steps.
For each dataset, we randomly select 1,000 samples, including 400 for training the alignment score predictor, 400 for calibration, and 200 samples as test data.

\paragraph{Results.}
The experimental results on TriviaQA are presented in Figure \ref{fig:llmtri}, while the results on CoQA are provided in Appendix F.3.
We empirically evaluate the FDP\(_t\) at different timesteps \(t = 50, 100, 150, 200\) over 100 replications. 
Unlike prior experiments, we examine the performance of OCS-ARC at different target FDRs.
Results show that our method controls the average FDP\(_t\) below each target FDR at every evaluated timestep. 
Moreover, our method achieves high Power\(_t\), indicating that it effectively selects most of the high-quality answers. 
Overall, OCS-ARC offers a model-agnostic candidate selection framework that can be effortlessly integrated with any black-box model, including LLMs.

\begin{figure*}
    \begin{subfigure}[b]{0.49\textwidth} 
        \centering
        \begin{subfigure}[b]{0.47\linewidth} 
            \includegraphics[width=\linewidth]{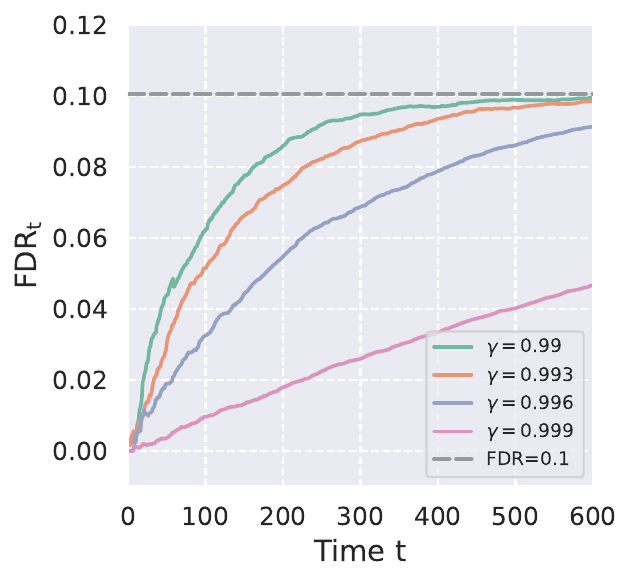}
        \end{subfigure}
        \hfill
        \begin{subfigure}[b]{0.47\linewidth} 
            \includegraphics[width=\linewidth]{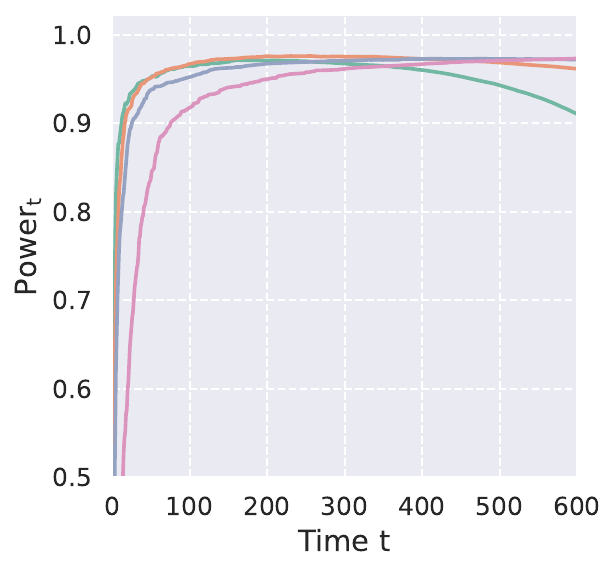}
        \end{subfigure}
        \caption{Decay Coefficient $r$}
        \label{fig:decay_coefficient}
    \end{subfigure}
    \hfill
    \begin{subfigure}[b]{0.49\textwidth} 
        \centering
        \begin{subfigure}[b]{0.47\linewidth} 
            \includegraphics[width=\linewidth]{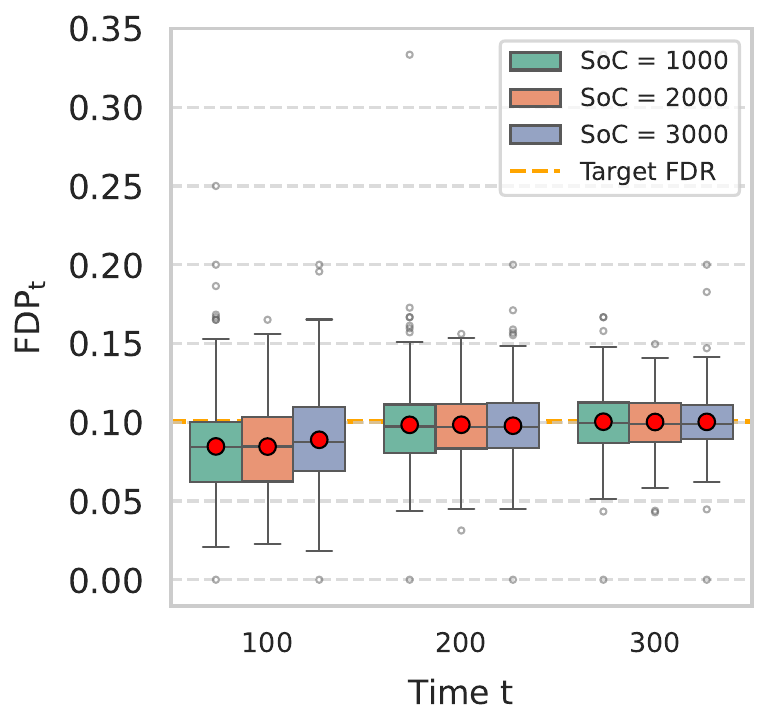}
        \end{subfigure}
        \hfill
        \begin{subfigure}[b]{0.47\linewidth} 
            \includegraphics[width=\linewidth]{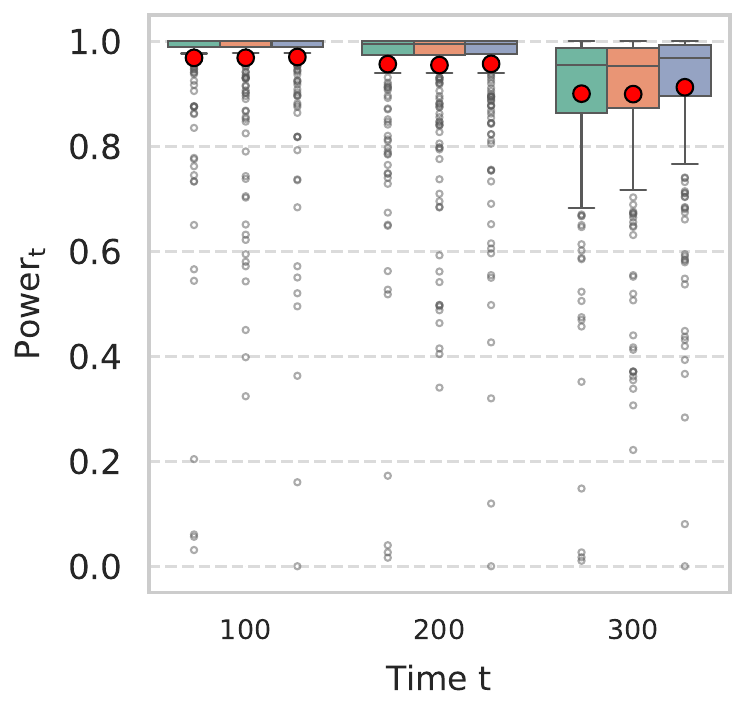}
        \end{subfigure}
        \caption{Size of Calibration Set (SoC)}
        \label{fig:cal_size}
    \end{subfigure}
\caption{\textbf{Results of parameter sensitivity analyses}, including (a) the effect of varying the decay coefficient $r$, and (b) the effect of changing the size of the calibration set. The target FDR is fixed at 0.10, and gradient boosting is used to fit the regression model. Parameter $r$ is chosen from the set \{0.99, 0.993, 0.996, 0.999\}, while SoC is varied over \{1000, 2000, 3000\}.}
    \label{fig:sensitive}
\end{figure*}

\subsection{Parameter Sensitivity Analysis}

Our approach employs a predefined decay coefficient \(r\) and requires a separate calibration dataset $\D_{\text{cal}}$.
In this section, we investigate the effect of \(r\) and the size of the calibration set (SoC) on the performance of our OCS-ARC.

\paragraph{Results.}
We set the target FDR at 0.1, employ gradient boosting for model fitting, and generate data based on Setting 1 from our main numerical experiments.
Figure~\ref{fig:decay_coefficient} illustrates the impact of varying the decay coefficient \( r \), selected from the set \{0.99, 0.993, 0.996, 0.999\}. 
The results show that our method consistently ensures valid FDR control across different \( r \). 
Moreover, a larger \( r \) (e.g., 0.999) typically leads to a more conservative approach at smaller \( t \), resulting in a relatively lower Power$_t$. 
In Figure~\ref{fig:cal_size}, we demonstrate the performance of our method under different calibration set sizes \{1000, 2000, 3000\}, at  \( t = 100, 200, 300 \).
We observe that a larger calibration set slightly reduces the variance of FDR and power, leading to more stable outcomes.
In general, our method is robust to changes in the calibration set size.
Overall, the results confirm the robustness of our method with respect to its parameters and the calibration set size.

\section{Related work}
Conformal inference \citep{papadopoulos2002inductive, vovk2005algorithmic} is an emerging technique for uncertainty quantification that constructs prediction intervals guaranteed to contain the true label with a predefined probability.
It has been applied across domains, including regression \citep{lei2014distribution, romano2019conformalized, zeng2025parametric}, classification \citep{angelopoulos2020uncertainty, huangconformal, xi2025does, zhou2025semi, liu2025spatial}, and online settings \citep{gibbs2021adaptive, bhatnagar2023improved, DBLP:conf/icml/AngelopoulosBB24, xi2025robust}.
Within the framework, several studies focus on controlling the FDR in predictive settings, known as \textit{conformalized multiple testing} \citep{bates2023testing, jin2023selection, hu2024two, zhang2025conformal}. 
This encompasses two-sample tests \citep{hu2024two} and multiple testing applications such as outlier detection \citep{zhang2022automs, bates2023testing, zhang2025conformal} and data selection/sampling \citep{jin2023model, jin2023selection}. 
A recent line of work extends this framework to selective conformal inference that constructs prediction intervals for a specific subset of test data, under both offline \cite{bao2024selective, jin2025confidence} and online settings \cite{bao2024cap, sale2025online}.
However, these approaches do not address the \textit{online candidate selection} problem, where selection decisions are irrevocable—a critical requirement in practical applications like recruitment screening.
In this work, we extend conformal selection to the online ARC setting to bridge this gap.

\section{Conclusion}
In this work, we propose OCS-ARC, an extension of conformal selection to the online ARC setting.
We provide theoretical guarantees for OCS-ARC, showing that the method controls the FDR at or below the nominal level at any timestep under both i.i.d. and exchangeable data assumptions.
Furthermore, we theoretically demonstrate that our approach can be easily extended to multivariate response settings.
Extensive experiments show that our method significantly improves the selection power over the baseline, while ensuring valid FDR control at each timestep.
We hope the insights of this work can enhance the applicability and impact of conformal selection across diverse scientific and industrial domains.
\bigskip
\bibliography{aaai2026}
\end{document}